\documentclass[runningheads]{llncs}
\usepackage[T1]{fontenc}
\usepackage{graphicx}
\usepackage{booktabs}
\usepackage{tcolorbox}
\usepackage{multirow}
\usepackage{float}

\usepackage{caption}

\usepackage{amsmath} 
\begin{document}
\title{How Reliable are LLMs for Reasoning on the Re-ranking task?}
%
%


\author{Nafis Tanveer Islam \and
Zhiming Zhao\inst{*}}
\authorrunning{Islam et al.}
%
\institute{
Multiscale Networked Systems (MNS) Group,
University of Amsterdam, 1098 XH, the Netherlands, \\ 
\email{Email: \{n.t.islam, z.zhao\} @uva.nl} \\
}
\maketitle              
%


\newcommand\blfootnote[1]{%
  \begingroup
  \renewcommand\thefootnote{}\footnote{#1}%
  \addtocounter{footnote}{-1}%
  \endgroup
}

\begin{abstract}
With the improving semantic understanding capability of Large Language Models (LLMs), they exhibit a greater awareness and alignment with human values, but this comes at the cost of transparency. Although promising results are achieved via experimental analysis, an in-depth understanding of the LLM's internal workings is unavoidable to comprehend the reasoning behind the re-ranking, which provides end users with an explanation that enables them to make an informed decision. Moreover, in newly developed systems with limited user engagement and insufficient ranking data, accurately re-ranking content remains a significant challenge. While various training methods affect the training of LLMs and generate inference, our analysis has found that some training methods exhibit better explainability than others, implying that an accurate semantic understanding has not been learned through all training methods; instead, abstract knowledge has been gained to optimize evaluation, which raises questions about the true reliability of LLMs. Therefore, in this work, we analyze how different training methods affect the semantic understanding of the re-ranking task in LLMs and investigate whether these models can generate more informed textual reasoning to overcome the challenges of transparency or LLMs and limited training data. To analyze the LLMs for re-ranking tasks, we utilize a relatively small ranking dataset from the environment and the Earth science domain to re-rank retrieved content. Furthermore, we also analyze the explainable information to see if the re-ranking can be reasoned using explainability.
\blfootnote{* Corresponding Author}

\end{abstract}

\section{Introduction}


Re-ranking is a fundamentally important task in retrieving content from various knowledge bases \cite{gao2024smlp4rec} \cite{li2023hamur}. Proper re-ranking enables users to access the most relevant information based on their query. In general, ranking or recommendation systems initially utilize responses through a text-similarity-based approach. For instance, Elasticsearch \cite{elasticsearch2025} uses BM25 \cite{robertson1994bm25} to generate a candidate list for the end user's query. Furthermore, re-ranking models are deployed to filter out more irrelevant content and provide users with the most relevant content. Current re-ranking models usually focus on the accuracy of the ranked candidate list based on user preferences \cite{liu2020personalized}. While accuracy in ranking is important, different people have different choices based on their queries and personal priorities. Therefore, a clear rationale and reasoning behind the recommendation or ranking can help the end user make a more informed decision and save time searching for content.

Collaborative Filtering (CF) \cite{schafer2007collaborative} attempts to reason based solely on tracking user behaviors, but it is one of the most widely used technologies for predicting user preferences. These methods utilize similar tastes among other users to predict items by leveraging search history and personal information to select the correct item for a user \cite{wang2018tem} for providing ranking or recommendations. However, these systems lack the appropriate item-wise reasoning, which should provide users with a logic behind the results. However, some system provides generic reasoning, including suggestions, because nearby people have also liked the same thing \cite{herlocker2000explaining} or based on similar search items \cite{sarwar2001item}. Ideally, an explanatory system should provide textual information regarding the decision made by the re-ranking system. Furthermore, this also increases transparency, effectiveness, and trustworthiness in a system \cite{li2025g}.

With the rapid advancements in Large Language Models (LLMs), particularly in their capabilities of generating textual reasoning and recommendation \cite{li2023survey} \cite{gao2024llm}, these models have increasingly demonstrated their ability to create natural language explanations. Recent developments in explainable re-ranking and recommendation approaches \cite{li2021personalized} \cite{li2021extra} emphasize providing users with coherent justifications underlying the ranking or recommendation outcomes. These methods can be broadly categorized as training-based approaches \cite{li2025g}, such as the RAFT-based methodology \cite{zhang2024raft}, which explicitly generates explanations alongside recommendations, collaborative filtering strategies \cite{li2021extra}, or attribution-based methods that identify tokens with significant contribution scores within recommendations \cite{he2015trirank}.

While these methods for personalized recommendations and re-ranking heavily depend on the availability of user data, typically, such datasets include personal information, user search history, interactions with neighboring users, or users with similar interests. However, obtaining sufficient data to build highly effective recommendation systems poses significant challenges, as users often hesitate to share personal information voluntarily, or at times, it is completely unavailable due to the development of an entirely new system. Moreover, regulatory bodies like the General Data Protection Regulation (GDPR) \cite{gdpr2016} impose stringent requirements on data collection, making it increasingly difficult to gather user data without explicit consent.

\begin{figure}[h]
    \centering
    \includegraphics[width=.6\linewidth]{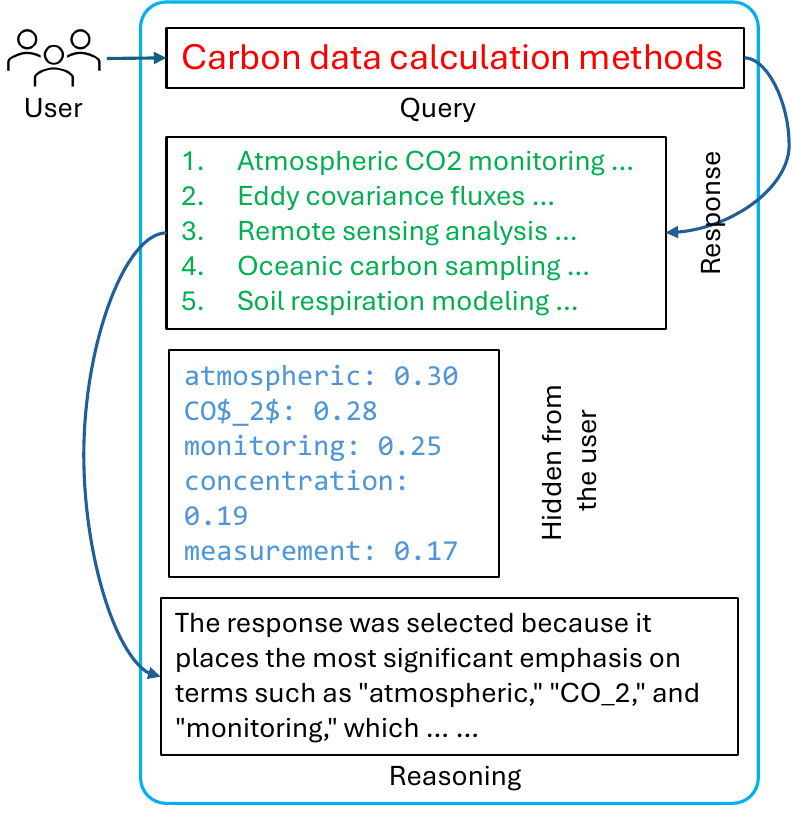}
    \caption{Contribution of our proposed work}
    \label{fig:impression}
\end{figure}

The scarcity of suitable ranking data and the lack of transparency among LLMs significantly complicate the task of delivering re-ranking or recommendations with clear reasoning. To address these challenges, we analyze various re-ranking techniques using large language models and propose an explainable approach that clarifies the rationale behind candidate rankings based solely on the query. For empirical evaluation, we utilize a small ranking dataset from ENVRI-Hub-Next \footnote{\url{https://envri.eu/envri-hubnext/}}, a specialized system designed for the environment and earth science domains that enables scientists to access relevant environmental datasets efficiently. We initially analyze different ranking methodologies using large language models. Finally, we use explainability techniques to transparently articulate the logical reasoning behind each recommendation, thereby enhancing user confidence and enabling informed decision-making. Figure \ref{fig:impression} illustrates how our system would interact in a real-world setting, where a user submits a query, receives a set of re-ranked lists or responses, and for each ranked list, we obtain the reasoning behind the item's selection.

In summary, our contributions are as follows:

\begin{itemize}
    \item We analyze and compare a set of re-ranking techniques using large language models. 

    \item We propose a SHAP explanation-based method powered by LLM to generate reasoning behind each recommendation.

    \item We also provide qualitative studies to analyze and support the textual reasoning of the ranked items.
\end{itemize}


\section{Related Work}


In this section, we discuss different methods of re-ranking and explainable re-ranking methodologies.

\subsection{Re-ranking}
Classical methods, such as BM25, dominated early research on re-ranking in information retrieval \cite{robertson2009probabilistic}, as well as learning-to-rank frameworks, including RankNet \cite{burges2010ranknet} and Coordinate Ascent \cite{burges2005learning}. These methods heavily rely on hand-crafted features, linear or ensemble models, and pairwise or listwise loss functions to optimize ranking effectiveness. The introduction of deep neural networks marked a significant shift in re-ranking research. Models such as DSSM \cite{huang2013learning}, DRMM \cite{guo2016deep}, KNRM \cite{xiong2017end}, and BERT-based architectures \cite{nogueira2019passage} brought significant improvements in retrieval and re-ranking quality by leveraging dense vector representations and contextual language modeling. The adoption of LLMs for general-purpose tasks has brought considerable advancement in ranking \cite{zhu2023large} \cite{liu2025llm4ranking}. Therefore, re-ranking methodologies have emerged as an essential intermediary task within recommender systems \cite{carraro2024enhancing} within LLM-based systems. Some studies leverage LLMs for zero-shot or few-shot re-ranking, prompt-based ranking, and generation augmented scoring \cite{gao2024llm}. Notably, LLMs can be combined with classical retrieval pipelines, providing semantic re-ranking and context-sensitive explanations for retrieved items \cite{liu2025llm4ranking}. Furthermore, recent works have demonstrated significant advancements in LLMs for re-ranking tasks with shorter contexts compared to longer contexts \cite{10.1145/3678004}.

\subsection{Explainable Re-ranking}

Prior approaches relied on heuristic and model-based explanations, such as content-based or collaborative filtering signals \cite{schafer2007collaborative}, which often provided only limited insights into why particular items were recommended \cite{zhang2014explicit}. Model-agnostic methods, such as LIME \cite{ribeiro2016lime} and SHAP \cite{lundberg2017shap}, were adapted for recommendation tasks to provide post-hoc explanations by attributing the output to key features or user-item interactions. More recent works have extended explainability to re-ranking scenarios, incorporating explainable AI into complex ranking pipelines \cite{ma2024xrec} \cite{peake2018explanation}. Approaches such as attribution-based explanation \cite{li2021extra} and multi-view learning \cite{li2021personalized} use feature importance or token-level attributions to produce interpretable rationales for re-ranking outputs. Large Language Models (LLMs) have also been utilized to generate natural language explanations \cite{yang2024maple} based on token attribution or user context, bridging the gap between model reasoning and human understanding. On the other hand, \cite{li2025g} employs a graph-based training method to generate explanations that originate from the training data.


Our analysis found that classical explainability methods have used attribution scores to comprehend token importance, while LLMs have been used in domain-specific settings for reasoning capabilities.  However, to the best of our knowledge, none of the previous works have utilized general explainable methods to generate textual explainability in re-ranking tasks. 

\section{Methodology}



In this section, we will initially outline the problem formulation for re-ranking tasks with text-based explainability.

\subsection{Problem Formulation}

The re-ranking task plays a vital role when we aim to provide a ranking of various queries while maintaining their information private, yet providing users with a valid reasoning for the ranking. Let us consider $U$ as the set of $n$ users where $U = \{u_1, u_1, u_3, ... , u_n\}$. Here, $U$ contains no personal information or identities about them except for their search preferences. $Q$ is the set of $k$ queries from $n$ users and $Q = \{q_1, _2, q_3, ... , q_k\}$. The set of $k$ queries may overlap between the $n$ users. For each of the queries in $Q$, by an user in $U$ will receive $m$ response items where the response items,  $I = \{i_1, i_2, i_3, ... , i_m\}$. To improve the recommendation, we propose a re-ranking methodology with a textual explainable method that provides users with a rationale behind each re-ranking and thereby explains the re-ranking process. Therefore, the users can make an educated decision when selecting a response item. Our analysis aims to generate refined response items of size $m^r (m^r < m)$, where $r$ refers to re-ranking. The goal of the re-ranking model is to enhance and provide users with a proper re-ranked item as shown in Equation \ref{eqn:1}.

\begin{equation}
\label{eqn:1}
    I^r = Rank(U, Q, I)
\end{equation}

Here, the re-ranked list of responses $I^r$ is a function of the user, their query, and the initial list of response items. Initially the ranking module provides a ranking score for each of the ranked item and then sorts the items. The items with the highest ranking score stays on top of the ranking and the lower scores goes at the bottom of the list $I^r$. After re-ranking, we provide the users with an explanation that focuses on the explainability scores. For a ranked item in $I^r$, we initially measure the top $K$ important keywords with an explainability score or attribution score $A_s$ based on Equation \ref{eqn:2}. 

\begin{equation}
\label{eqn:2}
    A_s = Ex(I^r, Q)
\end{equation}
Here, $Ex$ is a function that generates the explainability scores, which takes the re-ranked items $I^r$ and the query $Q$ as input and generates the explainability scores or attribution score $A_s$. Then, based on the $A_s$ for the query in $Q$, we use a prompt $P$ to provide a reasoning in a human-readable textual format $T$, which we formalize in Equation \ref{eqn:3}

\begin{equation}
\label{eqn:3}
    T = Reasoning(A_s, P, Q, I^r)
\end{equation}

Here, $T$ provides critical reasoning for why an item in $I^r$ is selected, which helps end users understand why the item was initially chosen for them.

\section{System Design}

\begin{figure}[t]
    \centering
    \includegraphics[width=1\linewidth]{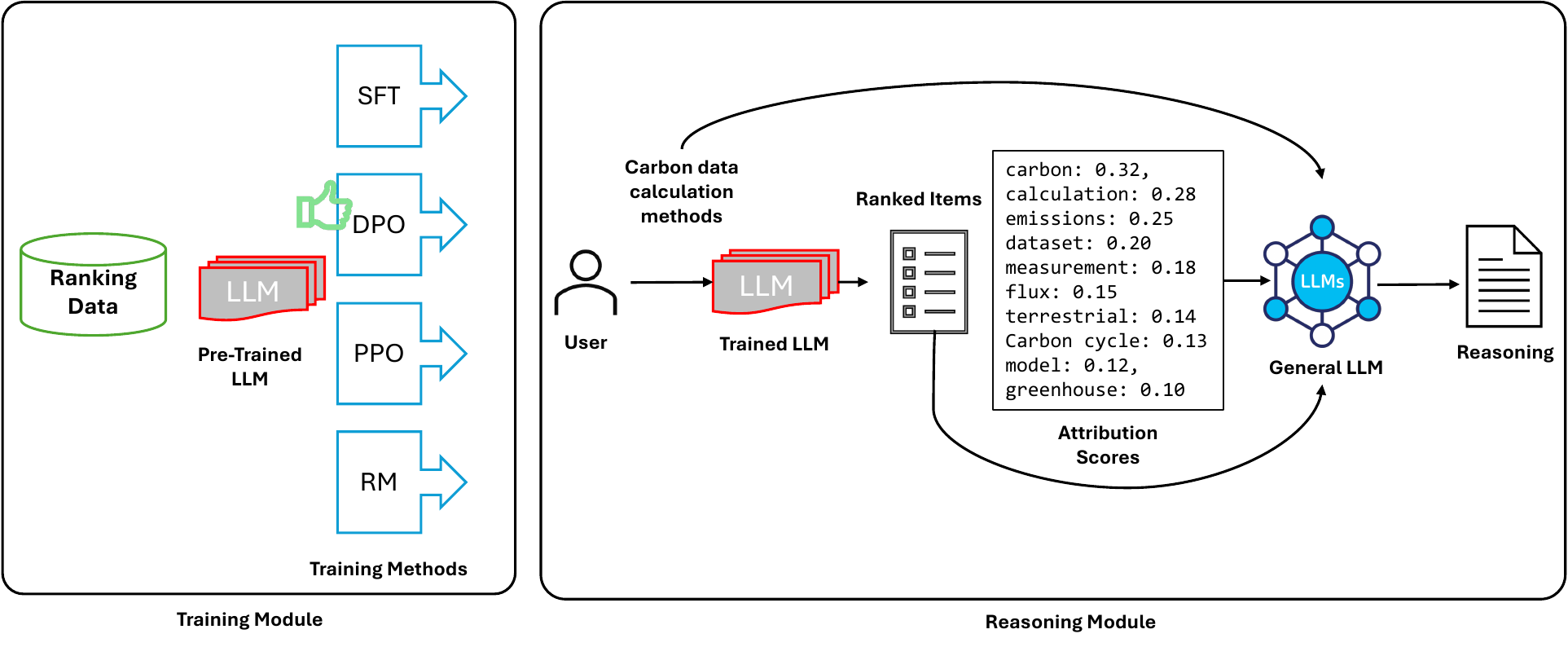}
    \caption{Overall Architecture of our Proposed System}
    \label{fig:arch}
\end{figure}

To develop our system, we propose an architecture that integrates various training methods for large language models, including supervised fine-tuning, direct policy optimization, proximal policy optimization, and reward modeling. Then, we use SHAP (SHapley Additive exPlanations) \cite{lundberg2017shap} values for an interpretable re-ranking task. Finally, the interpreted values or attribution scores are used by a general-purpose language model (LLM) to generate textual explanations. Our proposed architecture involves two core components, i) Training Module and ii) Reasoning Module. Figure \ref{fig:arch} shows an overall architecture of our proposed system.

\subsection{Training Module}
In this section, we discuss how we use different training modules for the re-ranking task.

\paragraph{\textbf{Re-ranking.}} To optimize LLM-generated re-rankings, we use various training methods. We use Supervised Fine-Tuning (SFT) to establish task understanding from labeled examples, while Reward Modeling captures user preferences for personalized optimization. Proximal Policy Optimization (PPO) fine-tunes rankings against this reward signal while constraining divergence from the base model. Direct Preference Optimization (DPO) is used when high-quality preference data is available, bypassing the need for a separate reward model. These different methodologies aim to re-rank the items. The training dataset consists of queries explicitly paired with corresponding item rankings. When each model is sufficiently trained, we retain only a single neuron from the last layer of the LLM and remove the rest, which will generate a single ranking score during inference. Based on the score, we create a re-ranking of the candidate items.

\subsection{Reasoning Module}
Here we discuss the components to use a trained LLM for re-ranking, then use SHAP explainability to generate attribution scores on the ranked item, and finally, to create natural text-based reasoning.

\begin{tcolorbox}[colback=black!5!white,colframe=green!75!black,title=LLM Explanation Generation Prompt]
\label{prompt}
You are given the following information:

\begin{itemize}
    \item A \textbf{user query}: \texttt{<Insert user query here>}
    \item The \textbf{top 10 tokens} from a response, each accompanied by its \textbf{attribution value} indicating its importance in ranking.
    \item The \textbf{response item}, including its URL and full text content:
    \begin{itemize}
        \item URL: \texttt{<Insert response URL here>}
        \item Summary Text: \texttt{<Insert response text here>}
    \end{itemize}
\end{itemize}

Using the attribution values of the top tokens, generate a concise, clear, and coherent explanation describing \textbf{why this response was selected} to answer the query. Focus on how the most important tokens contribute to the relevance and ranking of this response.

Your explanation should help a user understand the reasoning behind the ranking decision based on the key contributing tokens.
\captionof{figure}{Prompt template used for generating LLM explanations}

\end{tcolorbox}




\paragraph{\textbf{Interpretation with SHAP.}} To ensure interpretability and transparency in the re-ranking decisions made by the re-ranking process, we incorporate SHAP (SHapley Additive exPlanations) analysis. SHAP calculates the contribution of each query with the selected response item, determining how significantly each item's token impacts the query. The SHAP values quantify feature importance, enabling us to identify and understand precisely which characteristics influence the re-ranking decisions.

To calculate the importance of each token in the re-ranked contents, we initially select one of the ranked contents by the LLM. These ranked items include a URL with a summary of the contents in text, typically 5-6 lines long. Then, using SHAP values, we identify the top $k$ tokens from the summary text that have the highest correlation with the initial query $q$ based on an attribution value. Here, $A_s$ is the set of attribution scores for all tokens in $T$ of a response item $i$ where, $A_s \in \{a_1, a_2, ..., a_k\}$. After generating scores
For the top 10 tokens, we feed the token, along with its corresponding attribution score and the original query, to an LLM. We perform this for all the ranked contents of the LLMs.

\paragraph{\textbf{Natural Language Reasoning by LLM.}} Finally, the insights derived from SHAP values are translated into natural language explanations by a general-purpose LLM. This step bridges the gap between technical interpretations and intuitive human understanding. The LLM synthesizes the numeric SHAP explanations into coherent narratives, detailing the rationale behind each ranking decision.

To enhance the explanation generation process, we utilize a prompt to guide the LLM's processing of the input. The prompt for the LLM asks it to systematically reason through each explanation step-by-step, clarifying intermediate logical steps that contribute to the final ranking outcome. By explicitly outlining its reasoning process, the LLM provides deeper insights into the decision-making rationale, significantly enhancing the transparency and comprehensibility of the explanations. We follow the prompt shown in Figure 3 to generate natural language explanations by an LLM.


\section{Experiments and Discussions}

In this section, we will perform the necessary experiments on the dataset we collected from the ENVRI-Hub-Next hackathon. Our experimental setup is based on the following three research questions:

\begin{itemize}
    \item RQ1: How are different training methods effective at re-ranking tasks?
    \item RQ2: How do different LLMs compare to basic re-ranking methods?
    \item RQ3: How does adding a prompt with SHAP explainability enhance the textual reasoning behind the re-ranking process?
\end{itemize}


\subsection{Experimental Setup}
For all experiments, we randomly partition the dataset into training, validation, and testing subsets using an 80:10:10 split. Each model is trained for 8 epochs with a maximum input sequence length of 2048 tokens. We employ a learning rate of $2 \times 10^{-5}$ and a batch size of 4. For sequence generation tasks, we use a beam size of 4 and set the temperature to 0.5 to provide a balance between diversity and coherence. The training process is executed on a single NVIDIA A100 GPU equipped with 40 GB of memory. To optimize resource utilization and minimize memory overhead, we integrate DeepSpeed \cite{aminabadi2022deepspeed} with the large language models (LLMs) used in the following experiments. This setup allows the model to maintain approximately 35 GB per GPU, facilitating an efficient training environment with reduced inference latency. After training the model sufficiently, we use the SHAP explainability method to generate the attribution scores of each token in a response item and from our experimental analysis, we found that for the ranking tasks, on average the top 10 tokens with the highest attribution scores has the highest impact when they are fed to the LLM for generating textual explanations. Finally, we use a general-purpose LLM GPT-4o to develop the natural language explanations.

\subsection{Dataset Collection and Processing}
The dataset used in this study was collected during the ENVRI-Hub-Next Hackathon, where participants and experts from different research infrastructures \cite{farshidi2023knowledge} in the environmental and earth science domains were asked to query and evaluate the responses. Each user was provided with a URL to the ENVRI-Hub Knowledge Base, where they can submit a query and receive responses. Users searched for content relevant to their domain expertise and re-ranked the top 10 results. These top 10 results were initially retrieved using the BM25 algorithm. These expert-generated re-ranking annotations serve as the core supervision data for our analysis, enabling the evaluation and training of our re-ranking model. The experts ranked the items on a scale of 0-9, where 9 indicates that the content matches the user query highly and 0 indicates that the item does not relate to the query at all.

For supervised fine-tuning \cite{pareja2024unveiling} and reward modeling \cite{christiano2017preferences}, we can use the query, response item, and ranked value data directly for training. However, training with DPO \cite{rafailov2023dpo} or PPO \cite{schulman2017ppo} requires further processing. For DPO or PPO training, the training module attempts to compare positive responses with negative responses to optimize learning. Therefore, to prepare the training data for DPO and PPO, we split the ranking for a single query into two parts: the positive responses consist of the top 5 rankings, and the negative parts consist of the bottom 5 rankings given by an expert for a single query. Now, we create a pair from the contents of the positive response with the negative responses, thereby creating a sample set of 25 from a single query. In total, 94 participants responded. Therefore, we have 94 × 25 = 2,350 training data points for DPO and PPO training. However, for SFT and Reward Modeling, since we can use each of the 10 ranking datasets as training data, we have a total of 940 data points.

To support reproducibility, all relevant source code will be released upon the
publication of this work.


\subsection{Evaluation Metrics}

For our experiments, we use the metrics that determine the correctness of the retrieval and ranking task. For these tasks, we use metrics NDCG@K to assess the correctness of the ranking. Furthermore, we also utilize BERTScore, BLEU score, Rouge-L, and Cosine Similarity to determine how the ranked contents match with the initially ranked content by the experts.



\textbf{NDCG@K (Normalized Discounted Cumulative Gain).} extends beyond simple relevance by considering the position of each relevant item in the ranking, assigning higher importance to items appearing earlier. This prioritization makes NDCG@K (where $K=5, 10$) well-suited to reflect real-world scenarios where users focus more on top-ranked results. We follow a similar approach by \cite{gao2024llm} to calculate NDCG.

\textbf{BERTScore.} \cite{zhang2019bertscore} evaluates semantic similarity between the generated explanations and reference texts by leveraging contextual embeddings from pre-trained BERT models \cite{devlin-etal-2019-bert}. Instead of relying on exact matches, BERTScore measures token-level similarity for semantic alignment. For every query, we compute the BERTScore of the top-ranked result and report the average across all queries.

\begin{table}[t]
\centering
\caption{Training loss and top-1 accuracy for each optimization strategy across three model variants. Lower loss and higher accuracy indicate better performance.}
\begin{tabular}{@{}lcc|cc|cc@{}}
\toprule
 & \multicolumn{2}{c|}{\textbf{LLaMa}} & \multicolumn{2}{c|}{\textbf{Mistral}} & \multicolumn{2}{c}{\textbf{Phi-3}} \\
\textbf{Training Module} & \textbf{Loss} & \textbf{Acc.~(\%)} & \textbf{Loss} & \textbf{Acc.~(\%)} & \textbf{Loss} & \textbf{Acc.~(\%)} \\ \midrule
\textbf{SFT}            & 0.47 & 71.2 & 0.43 & 73.6 & 0.41 & 75.0 \\
\textbf{PPO}            & 0.44 & 73.8 & 0.40 & 76.9 & 0.38 & 78.3 \\
\textbf{DPO}            & \textbf{0.38} & \textbf{78.5} & \textbf{0.34} & \textbf{82.0} & \textbf{0.31} & \textbf{85.2} \\
\textbf{Reward Model}   & 0.42 & 75.1 & 0.37 & 79.4 & 0.35 & 81.0 \\ \bottomrule
\end{tabular}

\label{tab:training_metrics}
\end{table}

\textbf{BLEU Score.} \cite{papineni2002bleu} measures the n-gram overlap between generated text and reference outputs. As a reference-based metric, it produces a score ranging from 0 to 1.

\textbf{Rouge-L.} \cite{lin-2004-rouge} is similarly a reference-based metric that calculates the longest common subsequence (LCS) between the generated output and the reference text. The resulting score reflects both precision and recall, indicating structural similarity between generated and reference content.

\textbf{Cosine Similarity.} offers a complementary embedding-level metric that captures semantic similarity between generated and reference outputs, independent of exact token overlap. Given the summary text of a response item $D = {w_1, w_2, ..., w_m}$ and a ground truth response summary text $\hat{D} = {\hat{w}1, \hat{w}2, ..., \hat{w}n}$, we compute sentence embeddings via an encoder $E$, yielding $E(D) = {e{w_1}, e{w_2}, ..., e{w_m}}$ and $E(\hat{D}) = {e_{\hat{w}1}, e{\hat{w}2}, ..., e{\hat{w}_n}}$. We then aggregate token embeddings as:

\begin{equation}
e_D = \frac{1}{|D|} \sum_{i=1}^{m} e_{w_i}, \quad e_{\hat{D}} = \frac{1}{|\hat{D}|} \sum_{j=1}^{n} e_{\hat{w}_j}
\end{equation}

Finally, the semantic similarity score is calculated using cosine similarity:

\begin{equation}
sim(D, \hat{D}) = \frac{e_D \cdot e_{\hat{D}}^T}{|e_D| \cdot |e_{\hat{D}}|}
\end{equation}

\textbf{Loss.} We train the models with a pairwise preference loss. The model sees a positive and a negative content and is penalized whenever it scores the negative content higher.

\subsection{Discussions}
In this section, we discuss and analyze our experiments in relation to the three research questions we defined earlier.

\paragraph{\textbf{Answering RQ1:}} In answering RQ1, we perform a study to determine which is the best training method to re-rank contents. Based on the best training method, we employ the ranking method to train and test various large language models (LLMs) for the re-ranking task.

Table \ref{tab:training_metrics} illustrates the effect of the training modules on convergence quality. Our analysis shows that DPO consistently achieves the lowest loss and highest accuracy across all three LLMs, LLaMa, Mistral, and Phi-3. Then, the margin by which it outperforms the next-best method (PPO) widens with scale. For Model 3 (our 7-billion-parameter configuration), DPO delivers an additional 6.9-percentage-point accuracy gain over PPO, compared with a 4.7-point increase in Model 1. Proximal Policy Optimization (PPO) improves slightly over standard Supervised Fine-Tuning (SFT), confirming that reinforcement-learning objectives can align the ranking policy more closely with user intent. Overall, these results suggest that DPO is the preferred training method for the subsequent experiments.

\paragraph{\textbf{Answering RQ2:}} From RQ1, we determined that DPO shows the highest performance in the re-ranking task. Therefore, in answering RQ2, we use DPO as the training method to train five pre-trained large language models: LLaMa 3.1, LLaMa 7.1, Mistral 7B, Phi-3 3.8B, and Qwen 7B on the dataset. We updated the last layers of this model with a single neuron and removed the rest so that the model only produces a single value, which is the ranking value of the response item. The higher the ranking value, the more the model considers the item to be of higher quality. Then, based on the ranking value, we re-rank the items for each query and calculate the results. NDCG@k score reflects the ranking score. On the other hand, for the BERTScore, BLEU, and Rouge-L scores, we compare the ground-truth ranking of items with the rankings produced by different models. If the ranking of an item by a human and the trained LLM are in the same position, this score would be higher. Otherwise, they would be lower. Therefore, higher values of these metrics indicate that the re-ranking task by the model is better and more aligned with human judgment.

\begin{table}[t]
\centering
\caption{Evaluation of classical BM25 and five open-source LLMs on re-ranking task.}
\begin{tabular}{l|c|c|c|c|c|c|c}
\hline
\textbf{Model}   & \textbf{Param.} & \textbf{NDCG@5} & \textbf{NDCG@10} & \textbf{Cos.\ Sim.} & \textbf{BERTScore} & \textbf{BLEU} & \textbf{Rouge-L} \\ \hline
\textbf{BM25}    &  ––             & 0.48 & 0.45 & 0.62 & 0.67 & 0.22 & 0.28 \\
\textbf{LLaMa}   & 3.1 B           & 0.62 & 0.60 & 0.74 & 0.79 & 0.31 & 0.40 \\
\textbf{LLaMa}   & 7.1 B           & 0.71 & 0.69 & 0.82 & 0.86 & 0.39 & 0.48 \\
\textbf{Mistral} & 7 B             & 0.68 & 0.66 & 0.80 & 0.84 & 0.37 & 0.46 \\
\textbf{Phi-3}   & 3.8 B           & 0.60 & 0.58 & 0.72 & 0.77 & 0.30 & 0.38 \\
\textbf{Qwen}    & 7 B             & \textbf{0.73} & \textbf{0.70} & \textbf{0.83} & \textbf{0.87} & \textbf{0.41} & \textbf{0.50} \\ \hline
\end{tabular}

\label{tab:llm_ranking_results}
\end{table}

\begin{figure}[htb]
    \centering
    \includegraphics[width=1\linewidth]{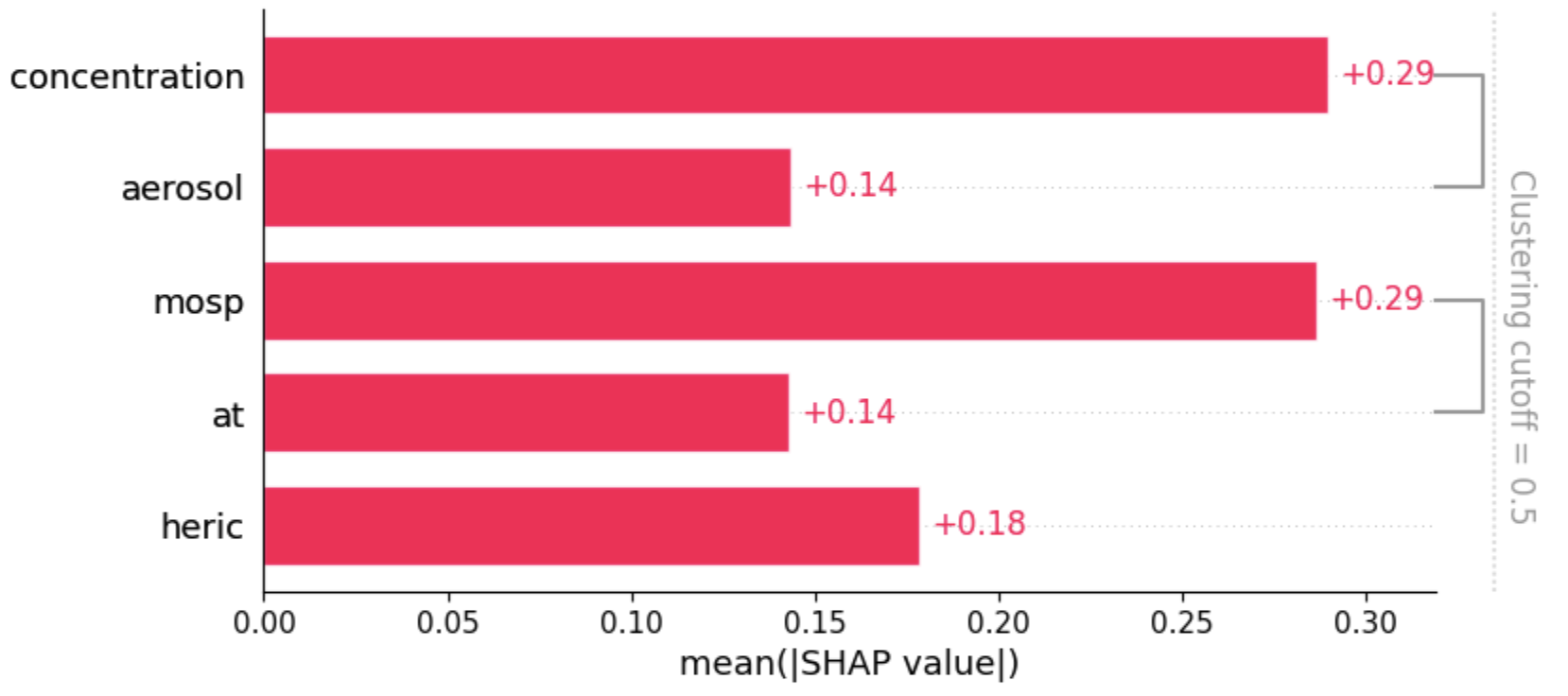}
    \caption{Top attribution scores of the selected item based on a query}
    \label{fig:attribute_score}
\end{figure}

The results from Table \ref{tab:llm_ranking_results} demonstrate that LLMs significantly outperform the BM25 baseline, underscoring the importance of large-scale language understanding in enhancing re-ranking effectiveness. Among the evaluated large language models, we observe a general trend that performance typically improves with an increasing number of parameters. Therefore, we see that the 7-billion-parameter LLaMa Qwen and Mistral consistently outperform the smaller 3.1B LLaMa or the 3.8B Phi across all metrics. Comparing the two leading 7-billion parameter models, LLaMa-7.1B and Qwen-7B, we observe that both models exhibit comparable performance. However, Qwen achieves the highest NDCG@5 scores on both test sets, as well as leading semantic and surface-level similarity metrics, including Cosine Similarity, BERTScore, BLEU, and Rouge-L. This superior performance likely reflects the benefits of Qwen’s bilingual pre-training and alignment strategies, which enhance both the relevance and the quality of explanations. The Mistral-7B model, while sharing the same parameter scale as LLaMa-7.1B, underperforms by approximately 0.03 higher in NDCG and 0.02 in BLEU. 

\begin{table}[!hb]
\centering
\caption{Environmental queries, top 10 attribution values of tokens in the response, and LLM-generated explanations (Part 1).}
\begin{tabular}{|p{2cm}|p{3cm}|p{7cm}|}
\hline
\textbf{Query} & \textbf{Attribution Value} & \textbf{Explanation} \\ \hline

Carbon data calculation methods & \begin{tabular}{@{}l@{}}
carbon: 0.32, \\
calculation: 0.28, \\
emissions: 0.25, \\
dataset: 0.20, \\
measurement: 0.18, \\
flux: 0.15, \\
terrestrial: 0.14, \\
carbon\_cycle: 0.13, \\
model: 0.12, \\
greenhouse: 0.10
\end{tabular} & 
\textbf{With $A_s$ :} The response is selected because it heavily emphasizes “carbon” and “calculation” related terms, highlighting datasets and measurement techniques relevant to carbon emissions. The presence of “flux” and “carbon\_cycle” indicates a focus on terrestrial carbon processes, which aligns well with the query’s intent to understand carbon data calculation.

\textbf{W/O $A_s$ :} The response provides comprehensive information on carbon data calculation methods, discussing various datasets and models related to carbon emissions and terrestrial processes.

\\ \hline

\end{tabular}
\label{tab:env_query_explanations_part1}
\end{table}

\paragraph{\textbf{Answering RQ3:}} In answering RQ3, we initially demonstrate quantitatively how the SHAP attribute scores are distributed across different tokens. Then, for each query and its response item, we generate the attribution score $A_s$. From the attribution scores, we take the tokens with the highest attribution score and then provide the $A_s$, prompt $P$, original query $q$, and the ranked item $i$. The prompt will guide the LLM to generate a textual response $T$, which will eventually explain the ranking.

\begin{table}[!ht]
\centering
\caption{Environmental queries, top 10 attribution values of tokens in the response, and LLM-generated explanations (Part 2).}
\begin{tabular}{|p{2cm}|p{3cm}|p{7cm}|}
\hline
\textbf{Query} & \textbf{Attribution Value} & \textbf{Explanation} \\ \hline

Atmospheric CO2 concentration trends & \begin{tabular}{@{}l@{}}
atmospheric: 0.33, \\
CO2: 0.31, \\
concentration: 0.29, \\
trends: 0.25, \\
dataset: 0.21, \\
measurement: 0.19, \\
ICOS: 0.17, \\
monitoring: 0.15, \\
greenhouse: 0.13, \\
emissions: 0.12
\end{tabular} & 
\textbf{With $A_s$ :} This response is selected because it focuses on “atmospheric CO2 concentration” and “trends,” which directly addresses the query. The mention of “ICOS” indicates the data source, reinforcing credibility. High attribution on “measurement” and “monitoring” suggests the response details robust observational methods.

\textbf{W/O $A_s$ :} The response presents data on atmospheric CO2 concentration trends, emphasizing measurements and monitoring from the ICOS network.

\\ \hline

Ocean temperature monitoring data & \begin{tabular}{@{}l@{}}
sea: 0.35, \\
temperature: 0.30, \\
dataset: 0.27, \\
monitoring: 0.22, \\
ocean: 0.20, \\
measurements: 0.18, \\
buoy: 0.15, \\
sensor: 0.13, \\
data: 0.12, \\
network: 0.10
\end{tabular} & 
\textbf{With $A_s$ :} The explanation highlights that the response contains key tokens such as “sea”, “temperature”, and “monitoring”, indicating that it provides relevant ocean temperature data. The attribution values show importance assigned to measurement technologies like “buoy” and “sensor”, confirming its relevance for precise oceanographic monitoring.

\textbf{W/O $A_s$ :} The response discusses ocean temperature monitoring datasets and measurement techniques involving buoys and sensors. 

\\ \hline

\end{tabular}
\label{tab:env_query_explanations_part2}
\end{table}

Figure \ref{fig:attribute_score} shows the response of the mean SHAP value for the query "\textit{Aerosol Concentration dataset}", indicating that the user is seeking a dataset related to aerosol concentration in a specific region. For this query, we initially select the top result with the highest ranking score from our trained LLaMa 7.1B model. The values in the $Y$ axis are some of the tokens with the highest values. We can see that the model assigns higher values to the token \textit{concentration}. Furthermore, we can also see that it has put a significant value into the split tokens at-mosp-heric, which is \textit{atmospheric} when we combine all the tokens. While this information is slightly tricky for a human to grasp because of the token \textit{atmospheric} split into three parts due to tokenization, when we use the prompt with the attribution values of the tokens, the query and the item as we have mentioned in equation \ref{eqn:3}, LLM generates a proper explanation for us. We used the following prompt (box \ref{prompt}) to create the answer.


Table \ref{tab:env_query_explanations_part1} and \ref{tab:env_query_explanations_part2} compare explanations generated by the LLM when provided with token attribution scores versus when such scores are withheld. When attribution values are available, the model explicitly references the most influential tokens that contributed to the ranking decision, yielding explanations that directly link token importance to relevance. Without attribution input, the LLM produces more general explanations based on semantic relevance inferred from the query and response content alone. This distinction highlights how integrating attribution information can improve transparency and user trust by providing grounded, token-level rationales for response selection.

\section{Conclusion}


In this work, we investigated the feasibility of generating logical and interpretable explanations for re-ranked items produced by large language models. We began with a comprehensive analysis of various training paradigms for re-ranking, ultimately identifying Direct Preference Optimization (DPO) as the most effective approach. Leveraging DPO, we systematically evaluated the performance of several state-of-the-art large language models (LLMs) on the re-ranking task, and we chose Qwen as the best for the re-ranking task based on our analysis. Subsequently, we generated token-level attribution scores for re-ranked items and employed a secondary LLM to produce explanations, using prompts that either included or excluded these attribution values. Our qualitative assessment indicates that explanations generated with access to attribution scores are markedly more coherent and relevant compared to those produced without such information. For future work, we aim to conduct both qualitative and quantitative analyses of explanation quality across a wide range of re-ranking use cases, incorporating human evaluations and automated explainability metrics to validate our approach.


\section*{Acknowledgment}
The work was made possible through funding from several European Union projects: ENVRI-Hub Next (101131141), EVERSE (101129744), BlueCloud-2026 (101094227), OSCARS (101129751). This research was partially funded by the Dutch Research Council (NWO) Large-Scale Research Infrastructures (LSRI) programme for the LTER-LIFE (http://www.lter-life.nl) infrastructure (grant 184.036.014).


%
%


\bibliographystyle{splncs04}
\bibliography{mybibliography}
%




\end{document}